\documentclass{article}
\usepackage{spconf,amsmath,amssymb,graphicx,booktabs,url,microtype,float}

\usepackage{enumitem}
\setlist{nosep,leftmargin=14pt}

\title{REST: Reliability-Guided Spatial Transcriptomics Supervision for Gene Expression Prediction from Histopathological H\&E-Stained Images}

\name{\begin{tabular}{c}
Lanfeng Zhong$^{1,2}$, Xiangyu Sun$^2$, Guixuan Xu$^3$, Rongzhao Zhang$^2$,\\
Jie Xu$^2$, Shaoting Zhang$^2$, Guotai Wang$^{1,2}$\thanks{Correspondence to Guotai Wang (guotai.wang@uestc.edu.cn)}
\end{tabular}}

\address{$^1$School of Mechanical and Electrical Engineering,\\
University of Electronic Science and Technology of China, Chengdu 611731, China\\
$^2$Shanghai Artificial Intelligence Laboratory, Shanghai 200232, China\\
$^3$Department of Pathology, Sichuan Provincial People's Hospital,\\
University of Electronic Science and Technology of China, Chengdu 610072, China}

\begin{document}
\ninept
\maketitle

\begin{abstract}
\looseness=-1
Histopathology is central to cancer diagnosis, but morphology in Hematoxylin and Eosin (H\&E)-stained images does not directly reveal the molecular states that characterize disease heterogeneity and inform tumor characterization, prognostic assessment, and biomarker discovery. Spatial Transcriptomics (ST) measures spatially resolved gene expression, providing molecular information within tissue architecture, yet differences in assay coverage and spot informativeness make paired H\&E--ST observations unequally reliable for representation learning. We propose \textbf{Reliability-Guided Spatial Transcriptomics (REST) supervision with Biological Program Bank (BPB)}, a program-structured and reliability-guided ST-supervised training framework for predicting molecular information from H\&E images. REST constructs BPB from curated signature, pathway, and cell-type gene sets to provide a shared structure for ST encoding, reliability estimation, and molecular reconstruction. Guided by BPB, Spot-Reliability Contrastive Alignment (SRCA) measures the reliability of each spot from assay-level program coverage and spot-level activity dispersion, and reweights H\&E--ST alignment accordingly. Biology-Informed Program Reconstruction (BIPR) further uses learnable program queries to reconstruct member-gene expression from histological morphology. REST's training corpus comprised 209 samples (407,475 spots across 16 organs); evaluation used 70 disjoint samples (230,056 spots across eight organs) for gene-expression, pathway-activity, and cell-signature prediction. REST achieved the highest Macro correlation among state-of-the-art ST-supervised training methods and outperformed existing pathology foundation models. Code is available at \url{https://github.com/lanfz2000/REST}.

\end{abstract}

\begin{keywords}
spatial transcriptomics, computational pathology, representation learning, contrastive learning, molecular prediction
\end{keywords}

\section{Introduction}
\label{sec:intro}

\begin{figure*}[t]
    \centering
    \includegraphics[page=1,width=0.98\textwidth]{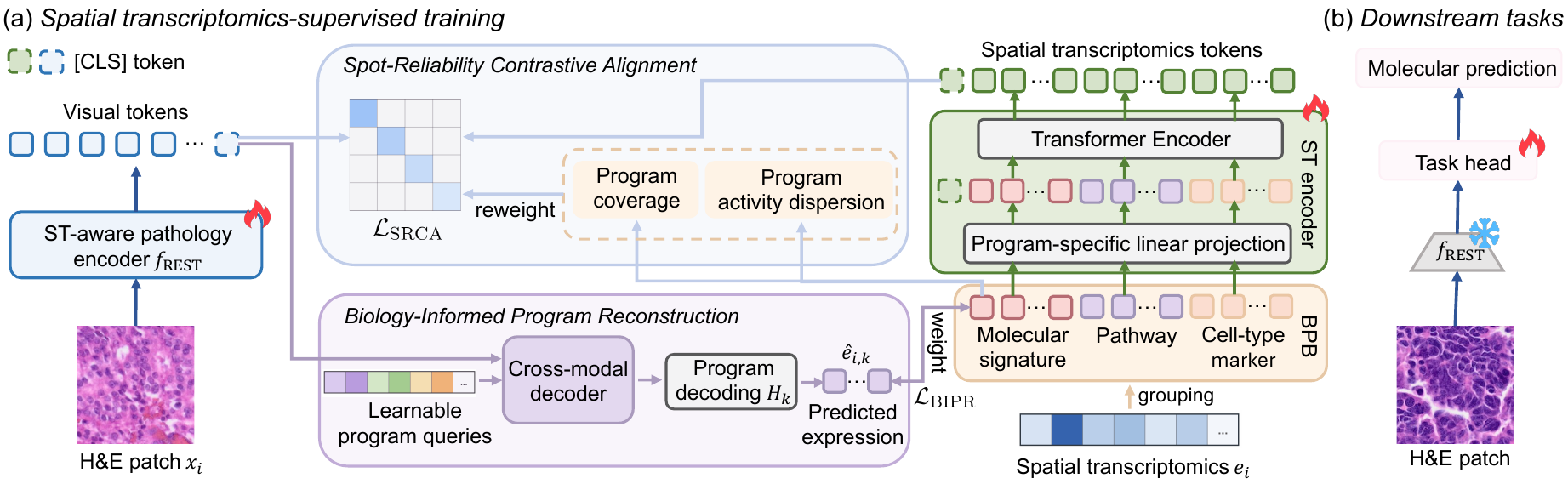}
    \caption{
    Overview of REST for obtaining ST-aware features from H\&E images for gene expression prediction.
    BPB: Biological Program Bank.
    }
    \label{fig:overview}
\end{figure*}

Hematoxylin and Eosin (H\&E)-stained images in histopathology are central to cancer diagnosis and treatment prediction, as they reveal tissue architecture, cellular morphology, and tumor organization~\cite{campanella2019weak}.
However, morphology alone does not directly expose the molecular states that characterize disease heterogeneity and inform tumor characterization, prognostic assessment, and biomarker discovery.
Spatial Transcriptomics (ST) measures spatially resolved gene expression, providing molecular information in the context of tissue architecture and linking local molecular variation to histological morphology~\cite{stahl2016spatial,vickovic2019hdst}.
However, ST requires dedicated tissue processing and sequencing and remains far less accessible than routine H\&E images.
These practical constraints motivate predicting local molecular profiles directly from H\&E images~\cite{schmauch2020he2rna,he2020stnet}.

Such prediction requires visual representations that connect local tissue morphology with molecular state.
Large-scale self-supervised and vision-language pretraining provides transferable histology encoders~\cite{wang2022ctranspath,wang2023retccl,chen2024uni,vorontsov2024virchow,huang2023plip,lu2024conch,zhou2025contrastive}, but these objectives do not explicitly capture local transcriptomic variation.
Paired H\&E--ST data provide molecular supervision for histology encoders, including through batch-agnostic contrastive learning for image recognition~\cite{nishimura2025batchagnostic}.

Histology--transcriptomics methods have explored bulk-level molecular modeling~\cite{schmauch2020he2rna,chen2022pancancer}, spot-wise regression~\cite{he2020stnet,jia2023thitogene}, spatial graph modeling~\cite{zeng2022hist2st}, and H\&E--ST contrastive learning~\cite{xie2023bleep}.
Among them, OmiCLIP represents each spot as a sequence of highly expressed genes and contrasts it with the paired H\&E patch~\cite{chen2025omiclip}, HAGE combines global and local alignment with hierarchical centroid constraints and gene prediction~\cite{dang2025hage}, STAMP augments symmetric contrastive learning with Hallmark-program gene reconstruction~\cite{zhou2026stamp}, and SEAL learns an ST latent space before aligning pathology with molecular embeddings~\cite{hemker2026seal}.
These studies highlight the importance of ST-supervised training objectives, but two limitations remain.

First, gene availability and molecular coverage vary across ST assays and samples, yet standard contrastive objectives weight matched H\&E--ST pairs equally.
Pairs from assays with limited program coverage or spots with weakly discriminative program profiles may therefore provide unreliable cross-modal supervision.
Second, directly reconstructing individual genes exposes the model to high-dimensional and sparsely observed gene-level noise.
Molecular signature, pathway, and cell-type marker collections provide an interpretable structure for organizing related genes~\cite{liberzon2015hallmark,kanehisa2023kegg,franzen2019panglaodb}, but the gene-type and pathway priors used by HAGE and STAMP~\cite{dang2025hage,zhou2026stamp} do not explicitly determine assay- and spot-dependent reliability weights for H\&E--ST alignment.
Effective ST supervision should therefore account for assay- and spot-dependent reliability while using biological structure throughout training.

\looseness=-1
To address these issues, we propose \textbf{Reliability-Guided Spatial Transcriptomics (REST) supervision with Biological Program Bank (BPB)}, a program-structured and reliability-guided training framework for learning molecularly informative histology representations from paired H\&E--ST data.
BPB integrates curated signature, pathway, and cell-type gene sets into a unified molecular structure, and provides the shared basis for two training mechanisms: Spot-Reliability Contrastive Alignment (SRCA) estimates the reliability of each spot from assay-level program coverage and spot-level activity dispersion, whereas Biology-Informed Program Reconstruction (BIPR) uses program-specific queries to reconstruct member-gene expression from H\&E features.
Across HEST-derived gene, pathway, and cell-signature tasks, REST achieves the highest Macro Pearson correlation among state-of-the-art ST-supervised training methods and outperforms existing pathology foundation models.
Our contributions are:
\begin{itemize}
    \item We propose REST with BPB, a unified bank of molecular signatures, pathways, and cell-type markers that guides ST encoding, reliability estimation, and reconstruction for molecularly informed histology representation learning.
    \item We introduce SRCA, which estimates spot reliability from assay-level program coverage and spot-level activity dispersion to reweight H\&E--ST contrastive alignment.
    \item We introduce BIPR, which uses learnable program queries to extract program-specific visual features for reconstructing member-gene expression.
\end{itemize}

\section{Method}
\label{sec:method}

\subsection{Problem Formulation and Overview}

Let \(x_i\) denote an H\&E patch and \(e_i\in\mathbb{R}^{G}\) the normalized expression profile of its paired ST spot over a unified vocabulary of \(G\) genes.
Each spot is paired with the H\&E patch cropped at the same tissue location.
ST is measured at discrete spots rather than over the whole slide image, and different assays cover different gene panels, so each profile is only partially observed.
We encode this with a binary gene measurement mask \(o_i\in\{0,1\}^{G}\), where \(o_{i,g}=1\) if gene \(g\) is measured by the assay that produced spot \(i\) and \(0\) otherwise.
We aim to train an ST-aware pathology encoder \(f_{\mathrm{REST}}\) for molecular prediction from H\&E.

As illustrated in Fig.~\ref{fig:overview}, REST first constructs a Biological Program Bank (BPB) and then trains \(f_{\mathrm{REST}}\) with two BPB-guided mechanisms.
SRCA uses BPB to estimate spot reliability for H\&E--ST alignment, while BIPR uses program-specific queries to reconstruct member-gene expression from histology.

\subsection{Biological Program Bank and ST Encoding}

As gene coverage varies across ST assays and individual counts are sparse, we encode each spot by biological programs rather than by individual genes.
Signature, pathway, and cell-type programs group related genes so that the ST encoder represents their joint activity rather than isolated measurements.
HAGE~\cite{dang2025hage} incorporates gene-type priors into gene-expression prediction, while STAMP~\cite{zhou2026stamp} uses Hallmark pathways for ST encoding, H\&E--ST alignment, and gene reconstruction.
REST extends program-based supervision with BPB, which spans multiple program types and shares their memberships across ST encoding, reliability estimation, and molecular reconstruction.
A useful BPB should cover genes measured by ST assays and span both molecular and cell-type programs, since an ST profile reflects both.
We therefore use three complementary curated sources: Hallmark~\cite{liberzon2015hallmark} molecular signatures, KEGG~\cite{kanehisa2023kegg} pathways, and PanglaoDB~\cite{franzen2019panglaodb} cell-type markers.
After standardizing gene symbols, we remove programs with fewer than three valid genes and retain overlapping memberships, since one gene may participate in multiple biological processes.
This yields BPB with \(K=387\) programs over a unified vocabulary of \(G=8{,}607\) genes, comprising 50 Hallmark, 186 KEGG, and 151 PanglaoDB-derived programs (Table~\ref{tab:program_bank}).
We encode BPB as a binary membership matrix \(M\in\{0,1\}^{K\times G}\), where \(M_{k,g}=1\) indicates that gene \(g\) belongs to program \(k\).

\begin{table}[t]
\centering
\caption{Composition of BPB. Note that a gene may belong to multiple programs and sources.}
\label{tab:program_bank}
{\setlength{\tabcolsep}{3pt}
\begin{tabular}{llrr}
\toprule
Dataset & Program type & Programs & Unique genes \\
\midrule
Hallmark~\cite{liberzon2015hallmark} & Molecular signature & 50 & 4,384 \\
KEGG~\cite{kanehisa2023kegg} & Pathway & 186 & 5,245 \\
PanglaoDB~\cite{franzen2019panglaodb} & Cell-type marker & 151 & 3,111 \\
\midrule
Union & -- & 387 & 8,607 \\
\bottomrule
\end{tabular}}
\end{table}

For each ST spot, raw gene counts are normalized by total library size, scaled to 10,000, and transformed with \(\log(1+x)\).
The resulting expression profile is mapped to the unified gene vocabulary.
Genes unavailable in a given sequencing assay are zero-filled and masked by \(o_i\), while measured zero expression remains a valid target.

We then encode each ST spot as one token per biological program.
Let \(\mathcal{G}_k=\{g:M_{k,g}=1\}\) be the genes in program \(k\).
For spot \(i\), let \(E_k\) denote the expression vector of genes in \(\mathcal{G}_k\) and \(M_k\) their gene measurement mask.
A program-specific projection \(W_k\in\mathbb{R}^{d\times|\mathcal{G}_k|}\) and bias \(b_k\in\mathbb{R}^{d}\) map the masked expression \(E_k\odot M_k\) to an ST token, where \(\odot\) denotes the element-wise product:
\begin{equation}
u_{i,k}
=
\operatorname{SELU}
\left(
W_k\,(E_k\odot M_k)+b_k
\right)
\end{equation}

The projected tokens are stacked as rows of \(U_i\in\mathbb{R}^{K\times d}\).
A Transformer encoder aggregates them through a learnable class (CLS) token, whose output is linearly projected and \(L_2\)-normalized to obtain \(z_i\in\mathbb{R}^{d}\).
This program-aware encoding makes signature, pathway, and cell-type structure explicit before cross-program aggregation.

\subsection{Spot-Reliability Contrastive Alignment}

ST supervision varies in reliability because sequencing assays differ in gene coverage and individual spots exhibit program profiles with different levels of informativeness.
Standard contrastive learning treats all matched H\&E--ST pairs equally, allowing pairs from assays with limited coverage or spots with weakly informative program profiles to contribute as strongly as reliable ones.
We therefore introduce Spot-Reliability Contrastive Alignment (SRCA), which assigns each spot a parameter-free reliability weight during H\&E--ST alignment.
Let \(P_i\in\mathbb{R}^{N_v\times d_v}\) denote the sequence of \(N_v\) visual tokens produced by \(f_{\mathrm{REST}}\) for patch \(x_i\).
We pool \(P_i\) and project it to obtain \(h_i=\operatorname{norm}(W_h\operatorname{pool}(P_i))\), where \(W_h\) is learned and \(\operatorname{norm}\) denotes \(L_2\) normalization; \(z_i\) is the paired ST embedding.
Their pairwise similarity is defined as \(s_{ij}=\tau\,h_i^\top z_j\), where \(\tau\) is a learned positive scale parameter.

We estimate pair reliability using two complementary properties: \textit{program coverage} and \textit{program activity dispersion}.
For program \(k\), the gene coverage and mean activity of spot \(i\) are computed as
\begin{align}
c_{ik} &= \frac{\sum_{g\in\mathcal G_k}o_{ig}}{|\mathcal G_k|}
\label{eq:program_coverage}\\
a_{ik} &= \frac{\sum_{g\in\mathcal G_k}o_{ig}e_{ig}}{\max\!\left(1,\sum_{g\in\mathcal G_k}o_{ig}\right)}
\label{eq:program_activity}
\end{align}

Let \(\mathcal K_i=\{k:c_{ik}>0\}\) denote the programs observed for spot \(i\), with \(|\mathcal K_i|=K_i\).
We summarize its overall molecular coverage as
\(q_i=\frac{1}{K_i}\sum_{k\in\mathcal K_i}c_{ik}\)
and its activity dispersion as
\(d_i=\sqrt{\frac{1}{K_i}\sum_{k\in\mathcal K_i}\left(a_{ik}-\bar a_i\right)^2}\),
where \(\bar a_i=\frac{1}{K_i}\sum_{k\in\mathcal K_i}a_{ik}\) is the mean program activity.
Higher \(q_i\) indicates broader assay coverage, whereas higher \(d_i\) indicates a more distinctive spot-level program profile.

\looseness=-1
We combine the two factors with square-root compression as \(r_i=\sqrt{d_i{q_i}}\) and normalize by the batch mean,
\(w_i=r_i/\bar r\), where \(\bar r=\frac{1}{B}\sum_{j=1}^{B}r_j\) and \(B\) is the batch size.
Thus, a pair receives a large reliability weight only when its assay provides sufficient gene coverage and its spot exhibits informative variation across biological programs.

Finally, SRCA applies these reliability weights to symmetric H\&E--ST contrastive learning, where the two terms are the H\&E-to-ST and ST-to-H\&E losses:
\begin{equation}
\begin{aligned}
\mathcal L_{\mathrm{SRCA}}
=
-\frac{1}{2\sum_i w_i}
\sum_i
w_i
\bigg[
&\log
\frac{\exp(s_{ii})}{\sum_{j=1}^{B}\exp(s_{ij})}\\
+\;&\log
\frac{\exp(s_{ii})}{\sum_{j=1}^{B}\exp(s_{ji})}
\bigg]
\end{aligned}
\end{equation}

\subsection{Biology-Informed Program Reconstruction}

SRCA aligns H\&E and ST representations at the spot level, but does not explicitly specify which molecular signals should be encoded in local visual features.
Direct reconstruction of thousands of individual genes is also sensitive to noisy and sparse gene measurements.
We therefore introduce Biology-Informed Program Reconstruction (BIPR), which organizes molecular supervision according to curated signature, pathway, and cell-type programs.

For each H\&E patch, BIPR operates on the visual token sequence \(P_i\) produced by \(f_{\mathrm{REST}}\).
We associate each biological program with a learnable query \(q_k\).
A cross-modal decoder $\mathcal{D}$ performs cross-attention~\cite{vaswani2017attention} from the program queries $Q$ to the projected visual tokens, so that each query retrieves the visual evidence most relevant to its program:
\begin{equation}
D_i=\mathcal{D}\!\left(Q,\,P_iW_v\right),
\qquad
Q=[q_1,\ldots,q_K]
\end{equation}
where \(W_v\) is a learned projection and the decoder applies self-attention over \(Q\) followed by cross-attention over \(P_iW_v\).
The output token \(D_{i,k}\in\mathbb{R}^{d}\) aggregates visual evidence for program \(k\).

\looseness=-1
Program decoding \(H_k:\mathbb{R}^{d}\rightarrow\mathbb{R}^{|\mathcal G_k|}\) maps \(D_{i,k}\) to predicted member-gene expression:
\(\hat e_{i,k}=H_k(D_{i,k})\in\mathbb{R}^{|\mathcal G_k|}\),
where \(\mathcal G_k=\{g:M_{k,g}=1\}\) contains the genes in program \(k\), and \(\hat e_{i,k,g}\) is the prediction for gene \(g\in\mathcal G_k\).

Because different sequencing assays measure different genes, only observed genes contribute to the reconstruction objective.
In addition, a gene may belong to multiple programs.
To avoid repeatedly over-weighting shared genes, we use inverse membership weights \(\alpha_{s,g}=1/n_{s,g}\), where \(n_{s,g}\) counts the programs of type \(s\) containing gene \(g\).
Let \(\mathcal P_s\) index the programs of type \(s\), and let
\(
\Omega_s=
\{(i,k,g):
k\in\mathcal P_s,\,
g\in\mathcal G_k,\,
o_{i,g}=1\}
\)
denote all valid reconstruction targets for program type
\(s\in\{\mathrm{sig},\mathrm{path},\mathrm{cell}\}\), corresponding to molecular signatures, pathways, and cell-type markers, respectively.
The BIPR objective averages the expression reconstruction losses equally across the three program types:
\begin{equation}
\mathcal L_{\mathrm{BIPR}}
=\frac{1}{3}
\sum_{s\in\{\mathrm{sig},\mathrm{path},\mathrm{cell}\}}
\frac{
\sum_{(i,k,g)\in\Omega_s}\alpha_{s,g}(\hat e_{i,k,g}-e_{i,g})^2
}{
\sum_{(i,k,g)\in\Omega_s}\alpha_{s,g}
}
\end{equation}

\subsection{Overall Training and Inference}

REST jointly optimizes SRCA and BIPR.
\begin{equation}
\mathcal L_{\mathrm{train}}=\mathcal L_{\mathrm{SRCA}}
+\lambda\mathcal L_{\mathrm{BIPR}}
\label{eq:training_objective}
\end{equation}
where \(\lambda\) is the BIPR training weight.
For each downstream task, we fit a ridge-regression head using frozen \(f_{\mathrm{REST}}\) features and ST-derived targets from the downstream training folds (Fig.~\ref{fig:overview}(b)).
At inference, the frozen encoder and fitted task head predict molecular targets from H\&E patches without ST measurements.

\section{Experiments}
\label{sec:experiments}

\setcounter{topnumber}{1}

\begin{table}[t]
\centering
\caption{Comparison with state-of-the-art methods. All entries report Pearson correlation (\%).}
\label{tab:main_results}
\begin{tabular}{lcccc}
\toprule
Method & Gene & Pathway & Cell & Macro \\
\midrule
\multicolumn{5}{l}{\textit{Without ST-supervised training}} \\
UNI2~\cite{chen2024uni} & 40.54 & 36.99 & 27.27 & 34.93 \\
H0-mini~\cite{filiot2025distilling} & 38.14 & 35.11 & 31.34 & 34.86 \\
Virchow2~\cite{zimmermann2024virchow2} & 37.43 & 38.14 & 28.82 & 34.80 \\
GigaPath~\cite{xu2024gigapath} & 38.29 & 36.95 & 27.51 & 34.25 \\
CONCHv1.5~\cite{ding2025titan} & 36.69 & 35.79 & 28.23 & 33.57 \\
CONCH~\cite{lu2024conch} & 36.11 & 33.69 & 27.87 & 32.56 \\
H-optimus-1~\cite{bioptimus2025hoptimus1} & 41.13 & 36.09 & 29.89 & 35.70 \\
\midrule
\multicolumn{5}{l}{\textit{With ST-supervised training (initialized from H-optimus-1)}} \\
STAMP~\cite{zhou2026stamp} & 40.41 & 40.65 & 33.67 & 38.24 \\
HAGE~\cite{dang2025hage} & 41.09 & 40.71 & 34.08 & 38.63 \\
OmiCLIP~\cite{chen2025omiclip} & 41.07 & 39.25 & 31.86 & 37.39 \\
SEAL~\cite{hemker2026seal} & \textbf{41.15} & 36.66 & 31.17 & 36.32 \\
REST & 41.14 & \textbf{42.39} & \textbf{35.66} & \textbf{39.73} \\
\bottomrule
\end{tabular}
\vspace{-1.2ex}
\end{table}

\subsection{Benchmark, Protocol, and Implementation Details}

We evaluated histology representations on HEST-derived gene-expression, pathway-activity, and cell-signature prediction tasks~\cite{jaume2024hest}.
The Gene task evaluated individual-gene expression, the Pathway task evaluated scores for 32 KEGG gene sets, and the Cell task evaluated scores for six cell-type marker sets.
We used 637,531 paired H\&E--ST spots from 279 public HEST samples acquired using Visium or Visium HD~\cite{jaume2024hest}.
The ST-supervision training corpus contained 209 samples (407,475 spots; 16 organs), and the downstream evaluation cohort contained 70 samples (230,056 spots; eight organs).
The two cohorts shared no samples or spots.
Performance was measured by Pearson correlation, which quantifies the linear association between predicted and ST-derived target values, with higher values indicating stronger positive agreement.
For Gene, Pearson \(r\) was computed per target gene across test spots and averaged over genes within each benchmark task; for Pathway and Cell, it was computed per target within each test sample and averaged over valid sample-target pairs with at least 20 spots and nonzero variance.
Macro Pearson correlation was the unweighted mean of the Gene, Pathway, and Cell correlations.

All ST-supervised methods were initialized from H-optimus-1~\cite{bioptimus2025hoptimus1}.
BPB comprised 387 programs covering 8,607 genes (Table~\ref{tab:program_bank}).
H-optimus-1 produces \(N_v=256\) visual tokens with dimension \(d_v=1536\).
The ST encoder, BIPR program queries, and contrastive embeddings use a common hidden dimension of \(d=256\).
Both the ST encoder and BIPR decoder contain two Transformer layers with eight attention heads and a dropout rate of 0.1.
To update the histology encoder efficiently, we apply LoRA to the query and value slices of the fused query--key--value projection in each attention layer, leaving the key slice frozen, with rank 8, scaling factor 16, and dropout 0.1.
Training was performed for 10 epochs with a global batch size of 256 using AdamW.
The learning rate was \(5\times10^{-4}\), with a weight decay of 0.2 and 500 warm-up steps.
The logit scale \(\tau\) was initialized to \(1/0.07\).
We used \(\lambda=0.1\) for the BIPR objective.

Within the evaluation cohort, we froze each histology encoder and fitted task-specific heads on the downstream training folds (Fig.~\ref{fig:overview}(b)).
All tasks used the same linear evaluation pipeline, following prior work~\cite{jaume2024hest,hemker2026seal}: feature standardization, principal component analysis to 256 dimensions, and a ridge-regression task head.
All preprocessing and regression parameters were fitted on the training folds only and then applied unchanged to the held-out test folds, without task-specific encoder fine-tuning.
Test predictions used H\&E patches alone; test ST measurements were used only to compute the evaluation metrics.

\subsection{Comparison with State-of-the-Art Methods}

Table~\ref{tab:main_results} reports two comparisons.
The upper section compares REST with pretrained pathology foundation models, including H-optimus-1~\cite{bioptimus2025hoptimus1}, Virchow2~\cite{zimmermann2024virchow2}, UNI2~\cite{chen2024uni}, H0-mini~\cite{filiot2025distilling}, GigaPath~\cite{xu2024gigapath}, CONCHv1.5~\cite{ding2025titan}, and CONCH~\cite{lu2024conch}, evaluated without ST-supervised training.
REST outperformed these pretrained encoders across all three tasks, with a Macro Pearson correlation of 39.73\%.
Relative to its H-optimus-1 initialization, REST improved Gene, Pathway, and Cell prediction by 0.01, 6.30, and 5.77 percentage points, respectively, resulting in a 4.03-point gain in Macro correlation.
This corresponds to an 11.29\% relative improvement in Macro correlation.
The larger gains on Pathway and Cell indicate that REST primarily strengthens representations of coordinated molecular programs and cell-type-associated states.

The lower section compares ST-supervised training methods, including STAMP~\cite{zhou2026stamp}, HAGE~\cite{dang2025hage}, OmiCLIP~\cite{chen2025omiclip}, and SEAL~\cite{hemker2026seal}, all initialized from the same H-optimus-1 backbone to isolate differences among training objectives.
REST achieved the highest Pathway, Cell, and Macro correlations and the second-best performance on gene prediction, trailing the best result by only 0.01 percentage points.
Compared with HAGE, the strongest competing method by Macro score, REST improved Gene, Pathway, Cell, and Macro correlation by 0.05, 1.68, 1.58, and 1.10 percentage points, respectively.
These controlled results show that the REST training objective uses paired ST supervision more effectively than competing objectives.

\begin{table}[t]
\centering
\caption{Ablation of REST under the H-optimus-1 backbone. All entries report Pearson correlation (\%).}
\label{tab:ablation}
\resizebox{\linewidth}{!}{
\begin{tabular}{lccccccc}
\toprule
Configuration & BPB & SRCA & BIPR & Gene & Pathway & Cell & Macro \\
\midrule
H-optimus-1 & -- & -- & -- & 41.13 & 36.09 & 29.89 & 35.70 \\
\midrule
InfoNCE & -- & -- & -- & 40.50 & 39.06 & 34.00 & 37.86 \\
InfoNCE & \checkmark & -- & -- & 40.90 & 39.81 & 35.12 & 38.61 \\
SRCA only & \checkmark & \checkmark & -- & 40.89 & 42.06 & 35.31 & 39.42 \\
REST & \checkmark & \checkmark & \checkmark & \textbf{41.14} & \textbf{42.39} & \textbf{35.66} & \textbf{39.73} \\
\bottomrule
\end{tabular}
}
\end{table}

\subsection{Ablation Study}

\looseness=-1
We evaluated the contribution of each REST training objective in Table~\ref{tab:ablation}.
Standard InfoNCE training achieved a Macro score of 37.86\%.
Adding BPB improved Gene, Pathway, Cell, and Macro performance by 0.40, 0.75, 1.12, and 0.75 percentage points, respectively.
With BPB in place, SRCA further improved Pathway, Cell, and Macro performance by 2.25, 0.19, and 0.81 percentage points, respectively, with a negligible 0.01-point decrease in Gene correlation.
SRCA uses BPB to measure assay-level coverage and spot-level activity dispersion, indicating that BPB-guided reliability weighting primarily benefits structured molecular targets.
Adding BIPR increased Gene, Pathway, Cell, and Macro performance by 0.25, 0.33, 0.35, and 0.31 percentage points, respectively, raising the Macro score to 39.73\%.
This consistent improvement supports using the same BPB to organize reconstruction targets in addition to its role in reliability estimation.

\section{Conclusion}
\label{sec:conclusion}

We proposed REST, an ST-supervised training framework that uses BPB to connect reliability-weighted H\&E--ST alignment with structured molecular reconstruction.
Evaluation on 230,056 spots from 70 samples showed that REST outperformed competing ST-supervised training methods in Macro correlation across gene, pathway, and cell-signature prediction.
Larger gains on pathway and cell-signature prediction support learning coordinated molecular features through program-structured supervision.
Scaling REST with larger, more diverse paired datasets could support foundation models for molecularly informed pathology.
Future work should evaluate prognosis and treatment-response prediction across institutions using independent cohorts with clinical outcome annotations, and extend REST to other spatial omics platforms.

\clearpage
\section{Compliance with Ethical Standards}

\looseness=-1
This retrospective study used publicly available human histology and spatial transcriptomics data from HEST~\cite{jaume2024hest}. No new samples were collected for this study.

\section{Acknowledgments}

This work was supported by the Natural Science Foundation of Sichuan Province under grant 2025ZNSFSC0455.

\end{document}